\documentclass[twoside]{IEEEtran}
\usepackage{cite}
\usepackage{amsmath,amssymb,amsfonts}
\usepackage{algorithmic}
\usepackage{graphicx}
\usepackage{algorithm,algorithmic}
\usepackage{hyperref}
\hypersetup{hidelinks=true}
\usepackage{textcomp}

\usepackage{multirow,array,hhline}
\usepackage{framed,multirow}
\usepackage{graphicx}
\usepackage{babel}
\usepackage{amsmath}
\usepackage{subcaption}
\usepackage{amssymb}
\usepackage{latexsym}
\usepackage{url}
\usepackage{xargs}
\usepackage{textcomp}

\def\BibTeX{{\rm B\kern-.05em{\sc i\kern-.025em b}\kern-.08em
		T\kern-.1667em\lower.7ex\hbox{E}\kern-.125emX}}

\begin{document}
	\bstctlcite{IEEEexample:BSTcontrol}
	
	\title{Two-Steps Neural Networks for an Automated Cerebrovascular Landmark Detection}
	\author{Rafic Nader, Vincent L'Allinec, Romain Bourcier and Florent Autrusseau
		\thanks{This work was supported by the French RHU-ANR project “eCAN” \# ANR-23-RHUS-0013 and INSERM CoPoC \# MAT-PI-22155-A-01}
		\thanks{Authors R.N, F.A and R.B are with the Institut du Thorax, Quai Moncousu, 44307 Nantes, France (e-mails: Rafic.Nader@univ-nantes.fr, Florent.Autrusseau@univ-nantes.fr, Romain.Bourcier@univ-nantes.fr), F.A. is also with the LTeN, Polytech, rue Ch. Pauc 44306 Nantes, France; author V.L. is with CHU Angers (Vincent.LAllinec@chu-angers.fr).}}

	\maketitle

	\begin{abstract}
		Intracranial aneurysms (ICA) commonly occur in specific segments of the Circle of Willis (CoW), primarily, onto thirteen major arterial bifurcations. An accurate detection of these critical landmarks is necessary for a prompt and efficient diagnosis. We introduce a fully automated landmark detection approach for CoW bifurcations using a two-step neural networks process. Initially, an object detection network identifies regions of interest (ROIs) proximal to the landmark locations. Subsequently, a modified U-Net with deep supervision is exploited to accurately locate the bifurcations. This two-step method reduces various problems, such as the missed detections caused by two landmarks being close to each other and having similar visual characteristics, especially when processing the complete MRA Time-of-Flight (TOF). Additionally, it accounts for the anatomical variability of the CoW, which affects the number of detectable landmarks per  scan. We assessed the effectiveness of our approach using two cerebral MRA datasets: our In-House dataset which had varying numbers of landmarks, and  a public dataset with standardized landmark configuration. Our experimental results demonstrate that our method achieves the highest level of performance on a bifurcation detection task.
	\end{abstract}


	
	
	
	\section{Introduction}
	\IEEEPARstart{T}{he} detection of cerebral vascular bifurcations landmarks is important for multiple clinical applications, including enhanced diagnostic precision, surgical planning, and customized therapeutic interventions. A precise identification of these bifurcation points enables accurate vascular mapping, which is essential for diagnosing cerebrovascular diseases such as intracranial aneurysms, arterial malformations, and stenosis \cite{van2015completeness,rinaldo2016relationship,kayembe1984}. Consistent identification of landmarks is important to monitor any vascular modification. Such a practice is critical to follow the evolution of various diseases and evaluate the effectiveness of therapies over time \cite{Boucherit2021,JEONG20141}. Developing an automated method to identify brain landmark locations could greatly assist in clinical applications, given the increasing workload and the intricate nature of the detection procedure required by radiologists.
	The localization of anatomical landmarks has been proposed across various imaging modalities and organs  ~\cite{wang2016benchmark,wu2017automatic,payer2019integrating}. Traditional approaches include rule-based methods~\cite{kim2017vertebrae,grau2001automatic}, statistical shape (SSMs)-based methods ~\cite{norajitra20163d,kafieh2007automatic}, or conventional machine learning algorithms ~\cite{oktay2016stratified,vstern2016local,urschler2018integrating}.
	All these methods depend on the quality of handcrafted features and become too complex to manage as image complexity increases.
	Recently, deep learning algorithms  have achieved a great success in medical image analysis including  anatomical landmark detection predominantly through two main approaches.
	The first strategy employs convolutional neural networks (CNNs) for coordinate regression \cite{lee2017cephalometric, wu2017automatic,zhang2017detecting,noothout2020deep,zeng2021cascaded}, where the models analyze local pixel-level patterns to directly predict the landmarks coordinates. Remarkable success was achieved using fully convolutional network (FCN) based on  heatmaps regression approaches \cite{zheng20153d,payer2016regressing,payer2019integrating,chen2019cephalometric,xu2021hip,oh2020deep,WANG2022,ao2023feature}. This technique constructs heatmaps that represent the probability distribution of landmarks at each pixel, thereby enhancing detection robustness against noise and improving localization accuracy through probabilistic rather than deterministic outputs. Unlike the first technique, which utilizes CNN for coordinate/heatmap regression, the second strategy leverages deep reinforcement learning ~\cite{ghesu2016artificial,alansary2019evaluating}. In this process, an agent continuously improves its method to precisely identify  landmarks in images. It achieves this by learning via a reward-based system that reinforces good identifications and penalizes mistakes. Nevertheless, the intricate nature of these multi-agent systems makes it difficult to identify many landmarks at the same time.
	
	Other researchers have adapted  object detection and image segmentation networks to perform landmark detection. For example,  \cite{qian2019cephanet} implemented Faster R-CNN integrated with a multitask loss function to identify landmarks. The authors in \cite{faster}  proposed a coarse-to-fine deep learning method for  detecting cranio-maxillofacial landmarks  from cone-beam computed tomography (CBCT) using Faster R-CNN. On the other hand, Zhang et. al \cite{runshi}  employed a semantic segmentation network for bone structure segmentation, followed by a top-down heatmap-based approach with unbiased heatmap encoding and a distribution-prior coordinate representation method to accurately perform landmark detection in craniomaxillofacial regions.


	
	In the context of vascular landmark detection, few works have been proposed \cite{Bogunovic2013,Robben2016,bilgel2013,gurobi2015,dunaas2017,wang2017}
	for bifurcation labeling.  The majority of these methods, primarily rely on graph matching or predefined segmentation techniques. These approaches necessitate significant preprocessing steps, including the semi-automatic extraction of the vascular centerline \cite{chow} or the assumption of an existing centerline-plus-diameter model. Consequently, these methods cannot be directly applied to raw imaging data. 
	In a recent work \cite{NADER2023102919}, we have proposed a deep learning approach for an automatic detection and classification of vascular bifurcations along the circle of Willis ; although this approach showed significant progress, it relies on the performance of a pre-segmentation step, possibly leading to a sub-optimal result. Moreover it uses local patches to perform bifurcation classification without regressing the intersection points between the merging/splitting branches. Recently, the authors in \cite{TAN2024102364} proposed a multi-task UNet based network which accomplishes landmark heatmap regression, vascular semantic segmentation and orientation field regression in order to detect 19 cerebral landmarks.
	\begin{figure}[!ht]
		\begin{centering}
			\includegraphics[width=0.56\columnwidth]{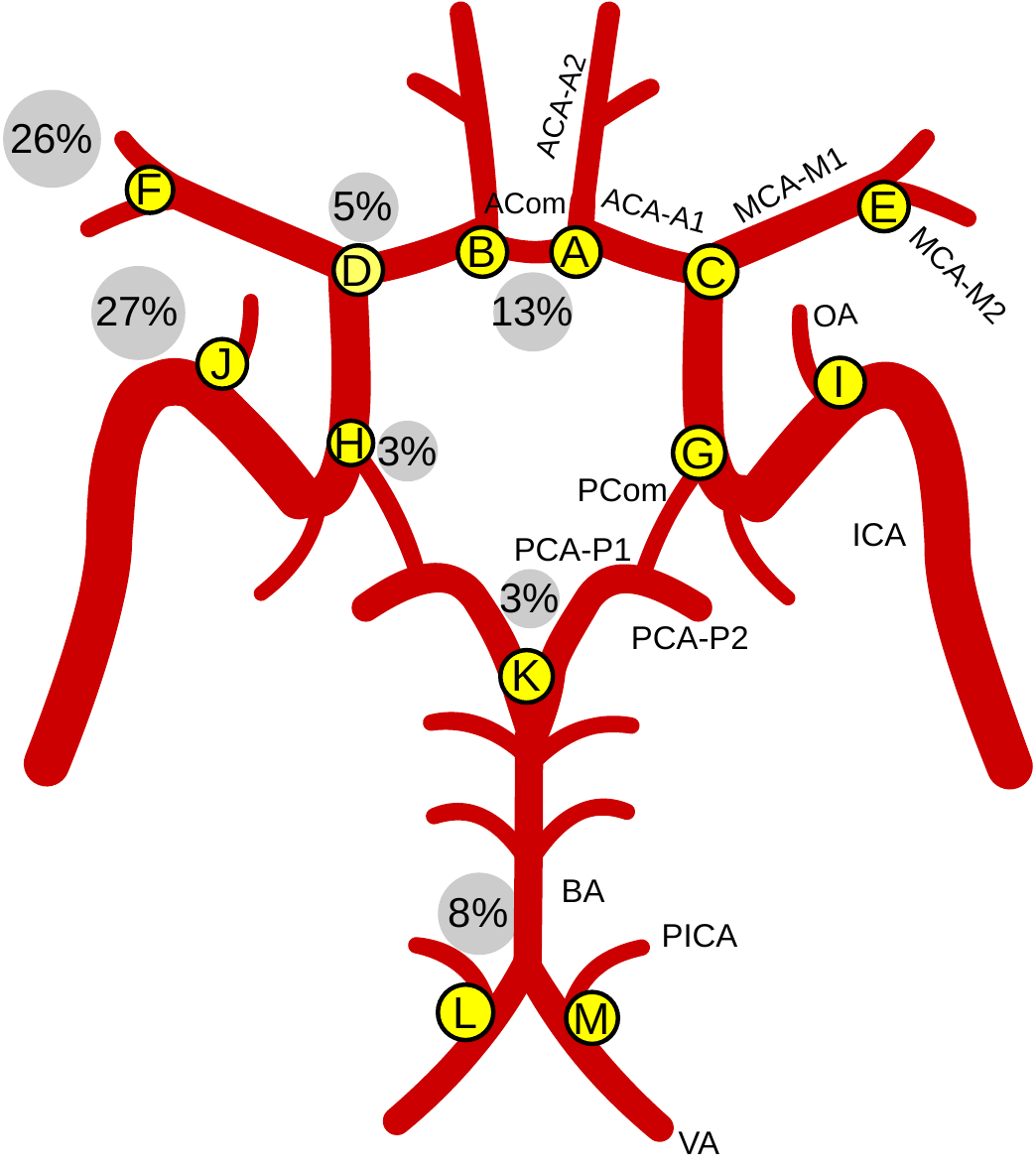}
			\par\end{centering}
		\caption{Schematic representation of the CoW.The yellow labels (from A to M) depict the particular bifurcations we are interested in for this study. The percentages within the gray discs represent the risk of aneurysm formation.
			\label{fig:SchematicWillis}}
	\end{figure}

	In this paper, we focus on 13 landmark points (bifurcations labeled A to M in Fig.~\ref{fig:SchematicWillis}) corresponding to the bifurcations representing the highest risk of aneurysm occurrence \cite{lazzaro2012,BROWN2014,Keedy2006}. This network exhibits a variety of configurations, characterized by differences in the number, shape, and size of the constituent arteries \cite{iqbal2013comprehensive,csahin2018}. Identifying anatomical landmarks is difficult because of the inconsistent structure of the Circle of Willis. Missing segments or deformations of the vessels, can modify the local appearance around landmarks, resulting in ambiguity during their recognition. Additionally, the size of 3D medical images often exceeds the memory capacity of GPUs, necessitating either a resampling or the design of a patch-based method. Furthermore, the approximate locations of certain bifurcations can cause heatmap predictions  to have strong responses at several nearby positions, which may result in detection errors.
	Besides, the current techniques for identifying landmark bifurcations, as described in \cite{payer2019integrating,TAN2024102364}, have practical constraints. They generally presume that different subjects have the same number of landmarks. However, this assumption is invalid because of the previously mentioned anatomical alterations in the Circle of Willis, as well as disparities in imaging technology and acquisition procedures. To circumvent this problem,  we introduce a two-step neural networks~\cite{faster} approach that is particularly effective in managing scenarios where the number of landmarks differs across images, enabling more flexible and precise detections.

	The rest of the paper is organized as follows. Section \ref{sec:MnM} describes the pipeline of the proposed method including the  the neural networks architectures and the evaluation protocol. In Section \ref{sec:ExpRes}, we present the experimental results. We conduct a comprehensive evaluation of the method's performance and compare it with  two of the state-of-the arts methods in landmark detection. Finally, Section~\ref{sec:Conclusion} concludes this work.

	\section{Materials and Methods}
	\label{sec:MnM}
\begin{figure*}[!ht]
		\begin{centering}
			\includegraphics[width=0.7\textwidth]{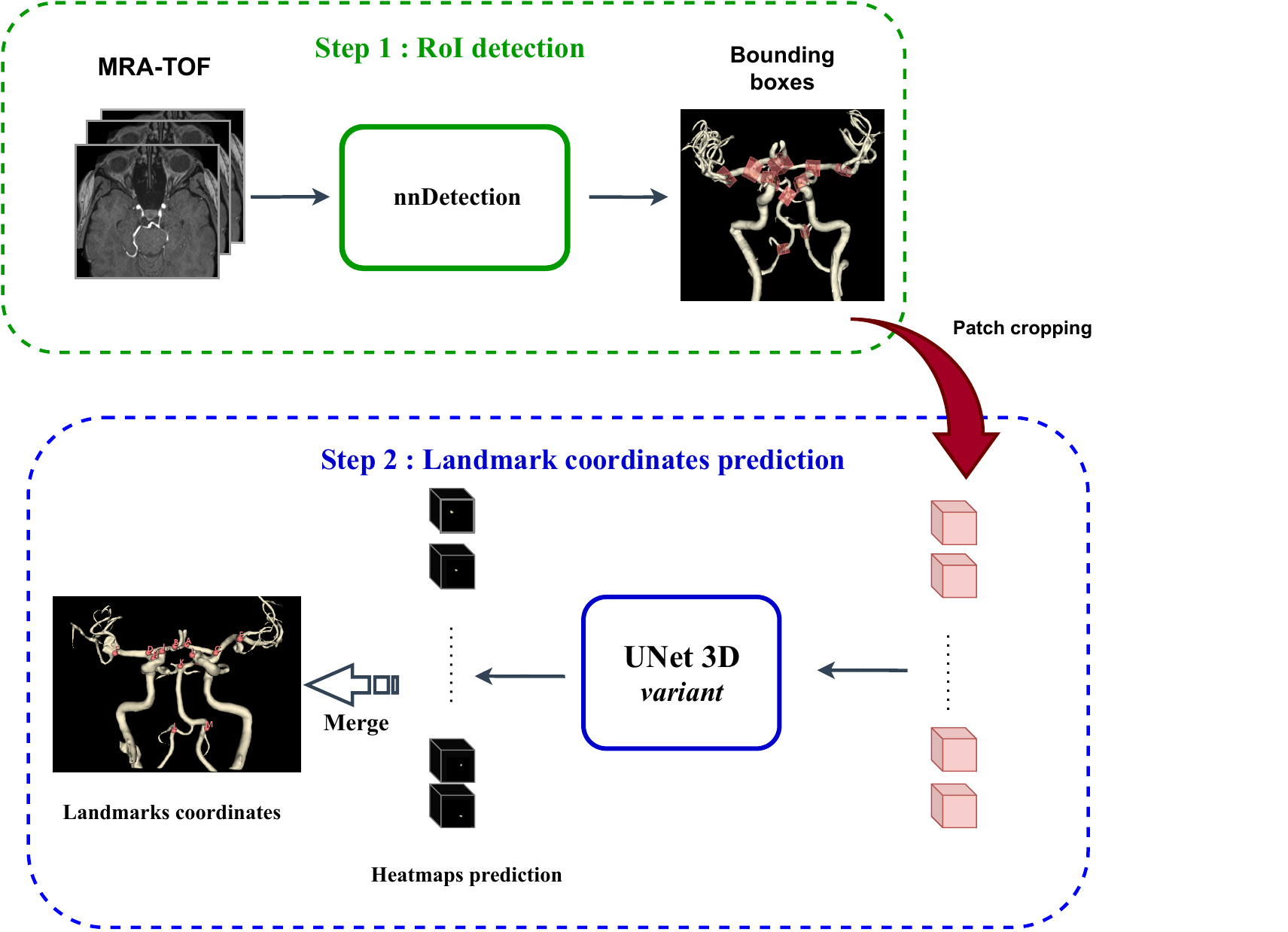}
			\par\end{centering}
		\caption{Schematic representation of the proposed landmark detection pipeline. The pipeline consists of 2 steps:  detection of region of interests (RoI) delimiting the landmarks (Step 1) and landmark coordinates prediction via a heatmap regression network (Step 2).\label{fig:pipeline}}
	\end{figure*}

	In this work, we propose a  two-step landmark prediction system for CoW bifurcations. Fig.~\ref{fig:pipeline} illustrates the workflow of the proposed pipeline. 
	In the first step, we formulate the problem as an object classification and bounding box regression problem. Landmarks are represented by fixed-size bounding boxes that are centered around them. The 3D nnDetection network predicts candidate
	regions of interest (ROIs) along with corresponding labels and scores.  In the second step, the precise coordinates of each landmark are determined  via heatmap regression using an encoder-decoder segmentation model. The second phase improves the initial detection, guaranteeing precise localization of landmarks and minimizing the identification of false positive detection.
	
	\subsection{Landmark position estimation}
	Let us first delve into the first process: the initial geographical localization estimation. 
	This step provides a coarse estimation of the bifurcation positioning (via a nnDetection), which will later on be refined (using a UNet).
	
	\subsubsection{Deep learning model}
	
\begin{figure*}[!ht]
		\begin{centering}
			\includegraphics[width=0.93\textwidth]{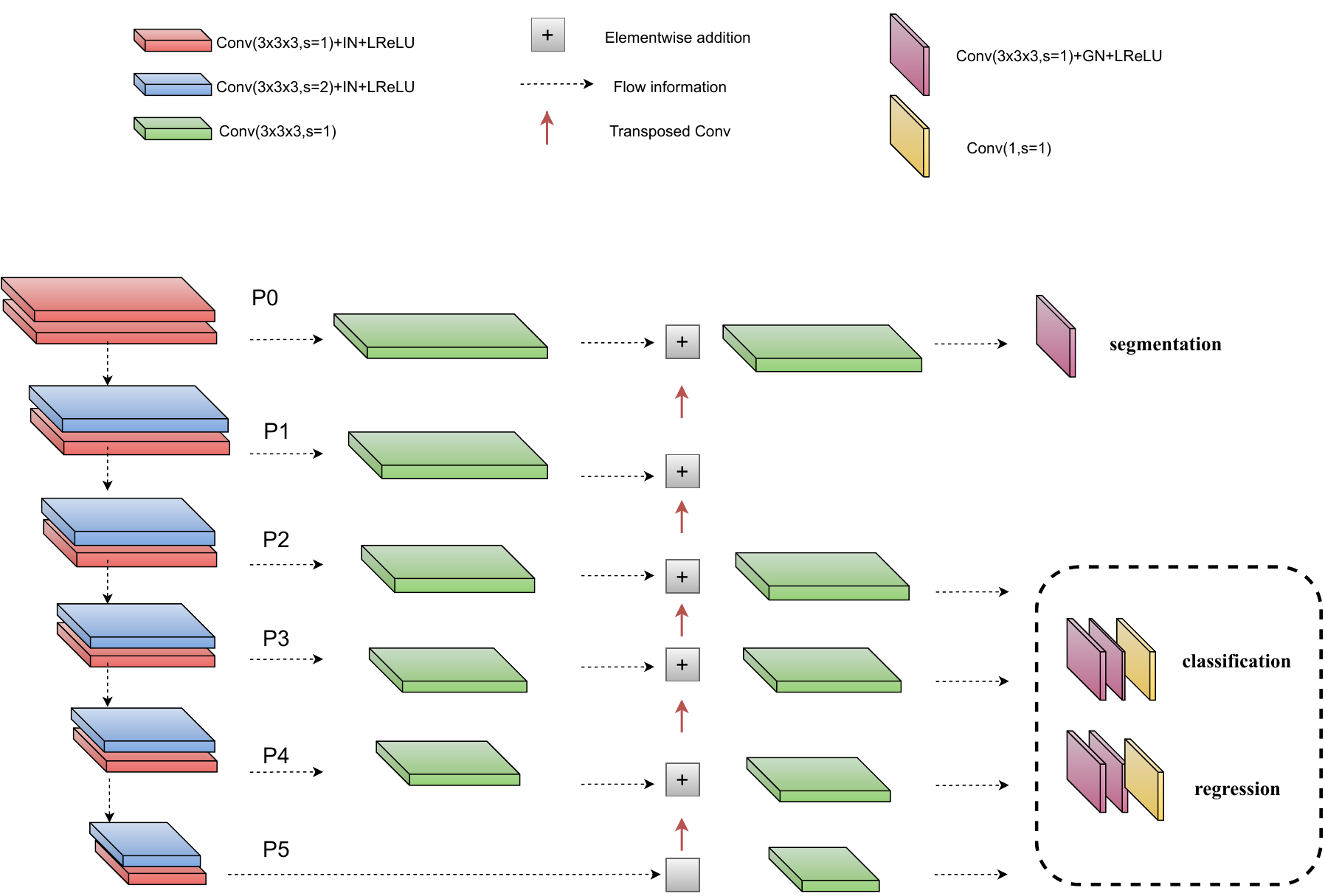}
			\par\end{centering}
		\caption{Retina U-Net  uses a backbone network (stacked convolutions) that extracts features at multiple scales. This backbone feeds into a feature pyramid network (FPN) that is essential for detecting objects of different sizes. The pyramid includes several layers (e.g., P0 to P5), where each layer represents a different scale of feature representation. The features maps (P2-P5) are used for the object detection and classification. Additional layers (P0-P1) are integrated into the FPN to support segmentation tasks. Similar to U-Net, skip connections are used for information flow.\label{fig:model}}
	\end{figure*}

	We used the state-of-the-art self-configuring framework for volumetric medical object detection, nnDetection ~\cite{baumgartner2021}, as our object detection network in this study. nnDetection  autonomously selects the optimal configuration for a given dataset by following a set of interdependent principles: 1) fixed parameters,  which remain constant regardless of the data (architecture, loss function, augmentation); 2) rule-based parameters which utilizes a set of heuristics based on the relevant properties of the training data (patch size, anchor optimization); and  3) empirical parameters, which are  optimized only during test time (IoU threshold for the NMS, model selection, etc..).
	nnDetection follows the topology of Retina U-Net \cite{jaeger2020retina} combining the detection capabilities of the popular RetinaNet detector with the segmentation strengths of the U-Net architecture. Figure \ref{fig:model} shows the topology of a Retina U-Net architecture.
	
	Although we have used the rule based parameters and the empirical parameters without modifications, we did adapt the augmentation parameters to fit our task. We deliberately avoid applying horizontal flipping due to the symmetric nature of our classes, which could introduce bias and lessen the detection and classification accuracy.
	\subsubsection{Ground-Truth  annotations}
	nnDetection  uses  a hybrid neural network model combining aspects of RetinaNet detector and U-Net. Indeed,  for the  training step, bounding boxes and segmentation annotations are necessary. In the development of nnDetection for a landmark detection task, we consider each landmark as an object to be located and classified within the image. Initially, fixed-size bounding boxes are established around each landmark point, which standardizes the detection area and simplifies the identification process. When detailed pixel-level annotations are available (In-House dataset), the segmentation is performed within these bounding boxes to precisely delineate arteries at the landmark. Conversely, where there are no pixel-level annotations available (public dataset), the bounding box is used as an alternative to the segmented arteries. This method, although less comprehensive, still enables accurate identification of landmarks by considering the enclosed area as the desired region, providing a practical alternative in cases where thorough segmentation data is unavailable \cite{baumgartner2021}.
	
	When approaching landmark detection as an object detection problem, it is reasonable to choose a uniform size for creating  ground truth bounding boxes. The chosen size must strike a balance by considering various aspects: the bounding box (BB) should be large enough to capture contextual information around the landmark, providing the network with relevant spatial information.  On the other hand, if the bounding box (BB) is very large, it may contain multiple landmarks, which might confuse the network during its classification step, and hence, decrease the localization accuracy. Considering these factors, we choose two distinct size aspects :

	\begin{enumerate}
		\item The BB size for landmarks E, F, I, J, K, L, and M is defined as 32 voxels. These landmarks are frequently separated from other bifurcations. A 32-sized bounding box, centered at the landmark coordinates, is used to enclose the branching structures related to the landmark. This gives the network enough context for an accurate identification.
		\item The ground truth BB sizes for landmarks A and B, C and G, and D and H are determined in such a way to encompass both landmarks simultaneously. These landmarks frequently appear in close proximity to one another, making it beneficial to establish a single bounding box that encompasses both landmarks, rather than separate boxes. In order to adequately include both landmarks, we begin by generating bounding boxes (BBs) with a size of 32 around each unique landmark. A unified bounding box, is then  generated to contain both of these initial boxes.

	\end{enumerate}
	
	Overall, we define 10 classes corresponding to 10 regions of interest.  These are determined as follows: \#R1 (encompasses landmarks $A$ and $B$), \#R2  (encompasses landmarks $C$ and $G$), \#R3  (encompasses landmarks $D$ and $H$), \#R4 (corresponds to landmark $E$), \#R5 (corresponds to landmark $F$), \#R6 (for landmark $I$), \#R7 (landmark $J$), \#R8  (landmark $K$), \#R9  (landmark $L$), and \#R10  (landmark $M$).
	
	\subsubsection{Training}
	The model undergoes training for a total of 60 epochs, with each epoch including 2500 batches. The training techniques are standardized using a batch size of 4 to ensure stability. For optimization, Stochastic Gradient Descent is employed with a Nesterov momentum term of 0.9. Additionally, a Poly learning rate schedule is implemented to vary the learning rate throughout epochs.
	The detection branch utilizes Binary Cross-Entropy as a loss function for object classification and Generalized IoU Loss for improving anchor box predictions. The segmentation branch is trained using the Cross-Entropy and Dice loss functions.
	\begin{figure}[t]
		\begin{centering}
			\includegraphics[width=0.90\columnwidth]{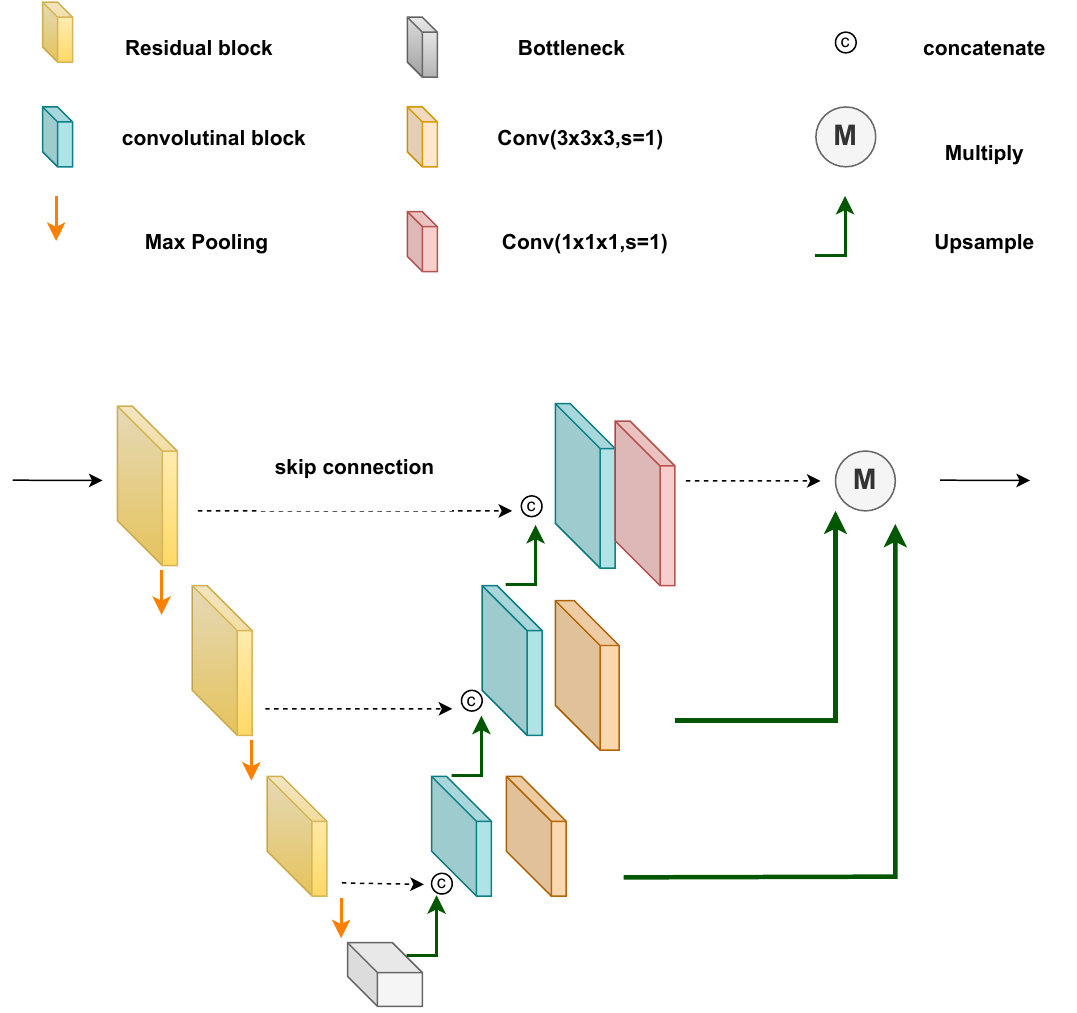}
			\par\end{centering}
		\caption{UNet variant for heatmap regression.\label{fig:model2}}
	\end{figure}
	
	\subsection{Landmarks position refinement}
	
	Once the approximate geometrical localization obtained through the previous network configuration, we need to pinpoint the target bifurcations; and especially for those bounding boxes enclosing two separate bifurcations of interest. To do so, we lake use of a second neural network as explained in the following. 
	
	\subsubsection{Neural Network}
	
	To find the exact location of each landmark within the ROI patch obtained from the first step, we used  UNet network \cite{Ronneberger2015} following the encoder-decoder architecture enhanced with deep supervision.  The detailed structure of the model is depicted in Fig. \ref{fig:model2}.
	The model takes as inputs 3D patches of size ($32\times32\times32$). Each level of the encoder consists of a residual block \cite{he2016deep} followed by a max pooling layer for downsampling. Each residual block is composed by two  3D convolution layers with  batch normalisation and ReLU activation. A residual connection is created between the input and output in order to alleviate  the vanishing gradient problem. For the decoder, each block contains an upsampling layer, concatenated with the features from the encoder  through skip connections and then processed using a convolutional block. Moreover, the deep supervision mechanism is incorporated by generating intermediate outputs at various stages in the upsampling path \cite{stawiaski2017multiscale,folle2019dilated,hilbert}. Each of these outputs is intended to predict the landmark heatmaps, which are then upscaled to match the final output dimensions.  In the final stages of the network, outputs from multiple levels are merged by multiplication, focusing on areas of agreement across the scales to enhance confidence and suppress noise in the predictions. To prevent over-fitting, we apply dropout. The last layer of each output is a $1 \times 1 \times 1$ convolutional layer with 13 channels, each one being a predicted heatmap for a specific landmark.
	For training the neural network, we used patches of size $32 \times 32 \times 32$  centered around the specific landmarks. The proposed network was trained for 500 epochs using Adam optimizer with $\beta_1=0.9$ and $\beta_2=0.999$ and a learning rate of 0.0001. 
	\subsubsection{Ground-Truth heatmaps}
	The ground-truth heatmaps are generated by applying a Gaussian distribution centered on the landmark coordinates.
	The intensity of the heatmap at any given voxel $x$ indicates the probability that this voxel represents the $i-th$ landmark. This probability decreases with the distance from the voxel to the landmark position $x_i$, modulated by the standard deviation $\sigma$, which determines the spatial dispersion of the landmark signal. The formulation for the heatmap associated with $N$ landmarks can be defined as:
	$$G_i(x) = \theta \times \exp\left(-\frac{1}{2} \left(\frac{x - x_i}{\sigma}\right)^2\right), \quad  i = 1, 2, \ldots, N,$$
	where $\theta$ is a scaling factor to avoid numerical instabilities due to small values of the Gaussian.
	We apply the mean squared error as a loss function between the predicted heatmaps $P_i$ and the Ground-Truth heatmap $G_i$.
	
	\subsubsection{Inference}
	During the test phase, the landmark coordinates are estimated by the argmax of the relevant heatmap determined by the label of the bounding box. For a BB encompassing two landmarks, the precise label for the heatmap is obtained by using a threshold that is determined by peak detection during the validation process. A landmark is considered missing if the peak intensity in a heatmap channel falls below this threshold.
	
	To summarize the entire inference procedure:
	\begin{itemize}
		\item In the first stage, bounding boxes and confidence scores are predicted using nnDetection. As multiple bounding boxes may be detected for the same label, only the region of interest (ROI) with the highest probability is retained for each label, and the rest are discarded.
		
		\item During the second stage, patches of size $32^3$ are cropped around the center of each predicted bounding box. These patches are processed through the UNet network to obtain the predicted heatmaps.
		
		\item  The argmax operation is performed on the associated channel(s) of the predicted heatmap for each label patch. Landmarks are kept if the pixel intensity exceeds a chosen threshold. As an illustration, in the case of category \#R1, we assess the highest intensity of a voxel in two heatmap channels (0 and 1) that correspond to landmarks A and B, respectively. These landmarks are grouped together inside the same region of interest (ROI). If the highest pixel intensity in either or both heatmaps exceeds the threshold, the coordinates of the corresponding landmarks are identified.
	\end{itemize}

	\section{Experimental results}
	\label{sec:ExpRes}
	
	\subsection{Datasets}
	In this study, we evaluate our method on two datasets with different vascular structures. We exploit our In-House MRA dataset with possible missing landmarks and a public MRA dataset with full landmarks.
	\subsubsection{In-House dataset}
	We have gathered an heterogeneous dataset of 145 images from the ICAN database \cite{NADER2023102919}. These images were issued from 28 French institutions and acquired from various MRI scanners. All the scans were spatially normalized (voxel spacing was set to $0.4 \times 0.4 \times 0.4$ $mm^3$). This dataset is split into a training data set of 97 images, and a test set of 48 images. For each volume, 13 bifurcations of interest (please refer to Fig. \ref{fig:SchematicWillis}) were annotated by a trained operator and validated by a neuroradiologist.  It is worth noting that the In-House dataset is not composed of a balanced amount of landmarks. This variability is caused by the MRA acquisition protocol where in certain cases the posterior cerebral tree is cut. Moreover, the CoW have naturally many different configurations among patients \cite{yang2023benchmarking} explained by aplasia (missing arteries) or hypoplasia (very thin arteries). We consider that when an artery is missing, the bifurcation is not present and thus, no landmark can be provided.
	
	\subsubsection{MRA public dataset}
	The public MRA dataset contains 104 scans taken from the UNC dataset (\url{https://public.kitware.com/WIKI/TUBETK/DATA}). The images were acquired by a Siemens Allegra head-only 3T MR system. Each image has a  voxel spacing of ($0.5\times 0.5\times 0.8 mm$). The annotation of 19 cerebral landmarks were labeled by the authors in \cite{TAN2024102364}. The annotation of the posterior landmarks (M and L) are absent for this dataset. We select the  remaining 11 landmarks from the annotated 19 landmarks. The public  dataset is only composed of scans with no missing bifurcations (the authors explicitly removed any scan having an incomplete number of landmarks).
	This dataset was split into 77 training images and 27 test images.
	
	\subsection{Evaluation}
	Following the configuration described in nnDetection, we used stratified fivefold cross-validation during the training of both networks (nnDetection and UNet) when applied to our in-House dataset. Briefly, the development set was partitioned into five folds, with four of the folds used to train the model, while the remaining fold served as the validation data to select the best-performing model through the epochs.  After the training, five different models were obtained, and the ensemble of these five models was used to predict the coordinates of the landmarks of the test set.
	When applied to the public MRA dataset, we used 61 images for training and 16 for validation for both networks. We test on the remaining 27 MRA-TOFs.
	To evaluate our method, we use  the two metrics used in the \textit{ISBI2015} Challenge: the Mean radial error (MRE) in mm and the Successful Detection Rate (SDR). The MRE computes the euclidean distance between the predicted locations and the ground truth locations of the landmarks. Whereas, SDR  measures the percentage of successful detection within a given neighborhood. In our case, we used three different neighborhood dimensions ($3~mm$, $4~mm$ and $5~mm$). For instance, in the following SDR-3 corresponds to a detection percentage when the predicted bifurcation center falls within a radius of $3~mm$ from the ground truth bifurcation center.
	
	For the In-House dataset, we also report the false positive landmarks and the false negative ones. A false positive (FP) is a landmark that was predicted but doesn't exist whereas a false negative (FN) is a true landmark that was not predicted.
	
	\subsection{Comparison with heatmap regression methods}
	To demonstrate the effectiveness of simultaneously employing object detection alongside heatmap regression compared to using only heatmap regression for the entire image, we evaluate our approach against two advanced deep learning methods for heatmap regression: SCN \cite{payer2019integrating} and FARNet \cite{ao2023feature}. The Spatial Configuration Network (SCN) integrates constraints on landmark relative positioning directly within the architecture, optimizing both local appearance and spatial configuration in a unified, end-to-end trainable framework.  Whereas, the method described in \cite{ao2023feature} introduces a novel deep learning architecture called FARNet, which is enhanced by a feature aggregation module for multi-scale feature fusion and a feature refinement module tailored to fine-tune high-resolution heatmap regression. We have re-implemented these two methods with their default settings and applied them to our datasets. The input images for both datasets were initially resampled to $0.6 \times 0.6 \times 0.8$ and then automatically cropped to dimensions of $192 \times 160 \times 160$ based on the mean landmark distributions.

	\subsection{Results}
	As previously explained, in this section, we provide a thorough evaluation of the proposed method, which we compared with two state-of-the-art methods of heatmap regression: SCN \cite{payer2019integrating} and FARNet \cite{ao2023feature}. 

	\subsubsection{In-House dataset}
	
\begin{figure}[!t]
		\begin{centering}
			\includegraphics[width=0.8\columnwidth]{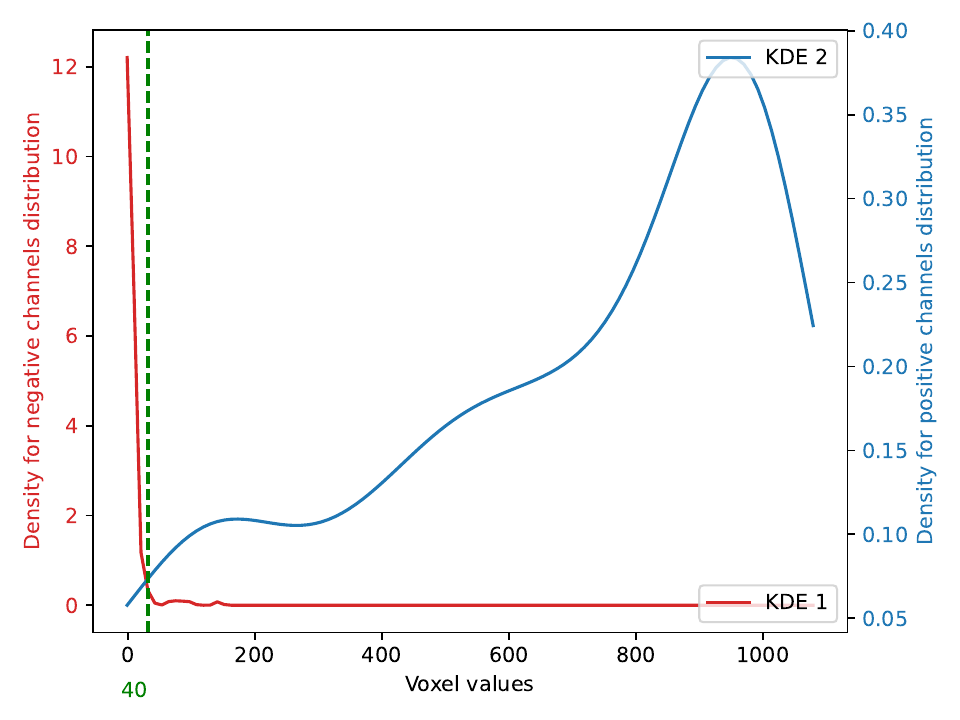}
			\par\end{centering}
		\caption{Voxel intensity distribution analysis of heatmaps in true positive channels vs other channels. We implement KDE (kernel density estimation) to convert the discrete voxel intensity values into continuous probability density functions for each type of heatmap and then we compute the first crossing point of the 2 plots.\label{fig:voxel}}
	\end{figure}

	Fig. \ref{fig:voxel} presents the distributions of maximum voxel intensities within heatmap predictions on the validation data of our dataset. For each identified landmark, we have collected the voxel intensities from two distinct groups : 1) the voxel values from the heatmap channel specifically associated with the landmark, and 2) the voxel values from all remaining channels. This approach allowed us to directly compare the specific responses of the true positive channel—those that are precisely correlated with the landmark with the non-specific responses exhibited by the other channels.  The first intersection point ($th=40$),  serve as a natural threshold. It can be applied to unseen test data to determine the landmark point present in a patch.
	
	\begin{table}[!ht]
		\caption{Detection comparison on the In-House MRA dataset, quantified using MRE and SDR for three neighborhood sizes ($3~mm$, $4~mm$ and $5~mm$).\label{tab1}}
		\centering{}%
		\begin{tabular}{|c|c|c|c|}
			\hline 
			\textbf{Metric/Method} & \textbf{FARNet} & \textbf{SCN} & \textbf{OURS} \tabularnewline
			\hline
			MRE (mm) & 2.73 & 2.38 & \textbf{1.97} \tabularnewline
			\hline 
			SDR($\%$)-3 & 72.17 & 81.73 & \textbf{82.31}\tabularnewline
			\hline 
			SDR($\%$)-4 & 82.36 & 88.69 & \textbf{91.24} \tabularnewline
			\hline 
			SDR($\%$)-5 & 90.08 & 92.69 & \textbf{95.27} \tabularnewline
			\hline
			\#FP &47 & 47& \textbf{9} \tabularnewline
			\hline
			\#FN&/&/&3\tabularnewline
			\hline
		\end{tabular}
	\end{table}
	For the In-House dataset, the results on the test set  are shown in Table \ref{tab1}. Our method achieves an average mean radial error of $1.97~mm$, which shows improvements of $0.39~mm$ over SCN and $0.76~mm$ over FARNet. While SCN performs similarly for small distance thresholds, our method gradually outperforms the others as the threshold increases, peaking at $95.27\%$ at $5~mm$. This improvement is especially remarkable in situations where landmarks have similar local appearances, a situation that often results in  localization errors with heatmap-based techniques like SCN and FARNet. These approaches primarily concentrate on local characteristics within each voxel, resulting in a significant number of incorrectly detected landmarks since they are not able to efficiently distinguish between significant and insignificant features. This can be seen if we compute the number of failures cases for each method.  Following \cite{payer2016regressing}, we consider that a prediction landmark is a failure case if the distance from predicted to ground-truth position is larger than $10~mm$. While we report 15 failure cases for SCN, 13 for FARNet, we have only 6 failure cases using our method. Consequently, our method not only reduces the mean radial error but also ensures a lower incidence of failure cases.

	\begin{figure*}[!t]
		\begin{centering}
			\includegraphics[width=0.72\textwidth]{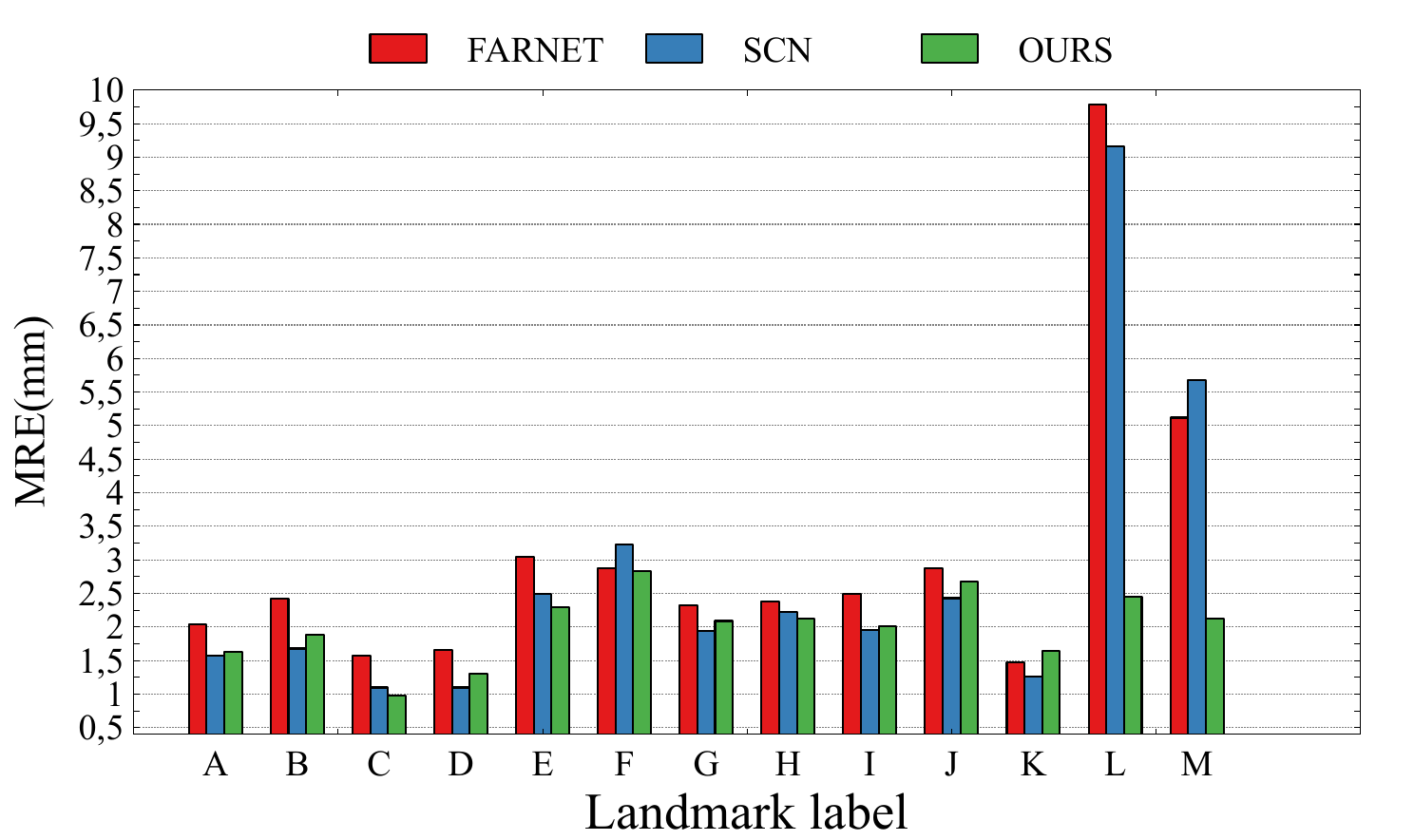}
			\par\end{centering}
		\caption{Mean radial errors (mm) for 13 landmark labels for the In-House dataset. \label{fig:bif}}
	\end{figure*}
	Moreover, our methodology includes an early phase of object detection, which, in our specific situation, results in the omission of 3 landmarks. Nevertheless, the significant occurrence of false positives (47 for both SCN and FARNet) compared to ours (9 FPs) highlights the constraints of heatmap-based approaches that solely rely on continuous heatmap predictions  and lack mechanisms to effectively handle missed landmarks.
	For applications requiring a balance between accuracy, low failure cases and low false positive detection, our method is distinctly advantageous. Furthermore, it can be used for datasets with a varying number of landmarks while the other methods have been designed for acquisitions comprising all the bifurcations of interest.

\begin{figure*}
		\centering
		\subfloat[GT]{%
			\includegraphics[width=0.28\linewidth]{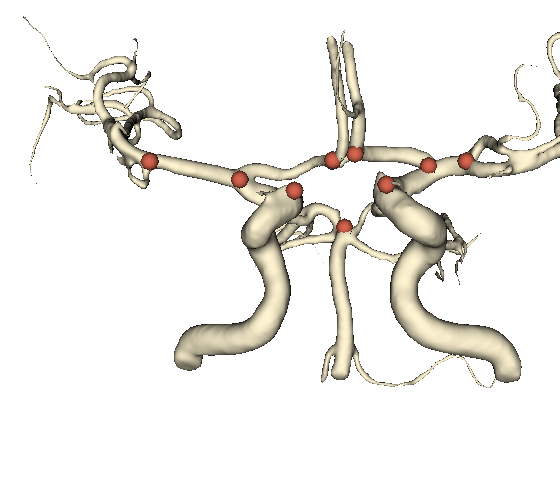}}
		\hfill
		\subfloat[GT]{%
			\includegraphics[width=0.28\linewidth]{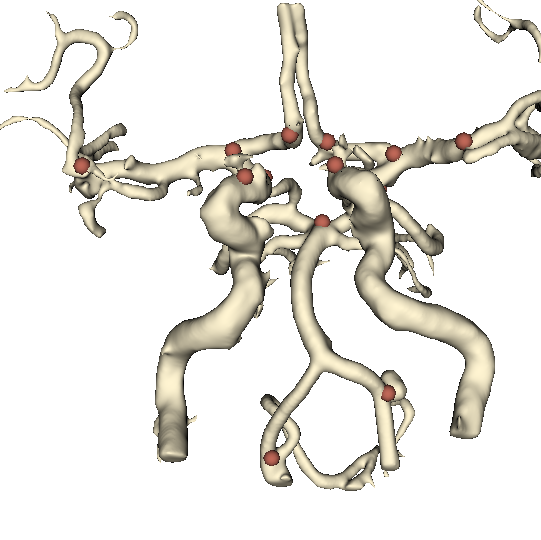}}
		\hfill
		\subfloat[GT]{%
			\includegraphics[width=0.28\linewidth]{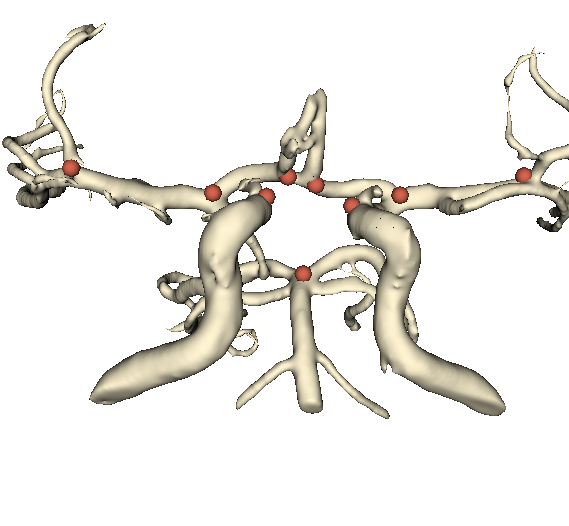}}
		
		\subfloat[SCN]{%
			\includegraphics[width=0.28\linewidth]{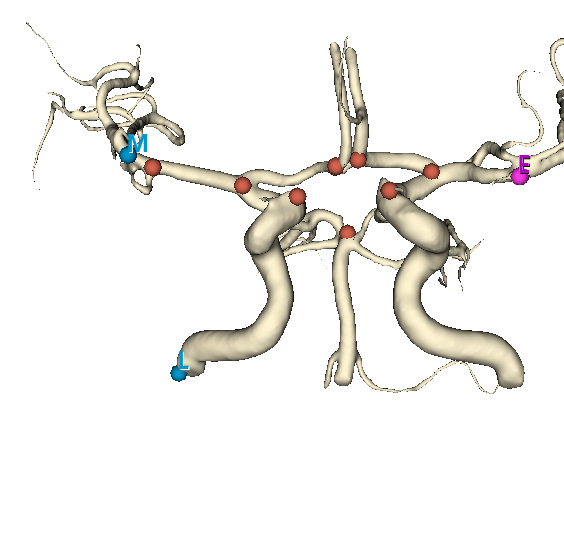}}
		\hfill
		\subfloat[SCN]{%
			\includegraphics[width=0.28\linewidth]{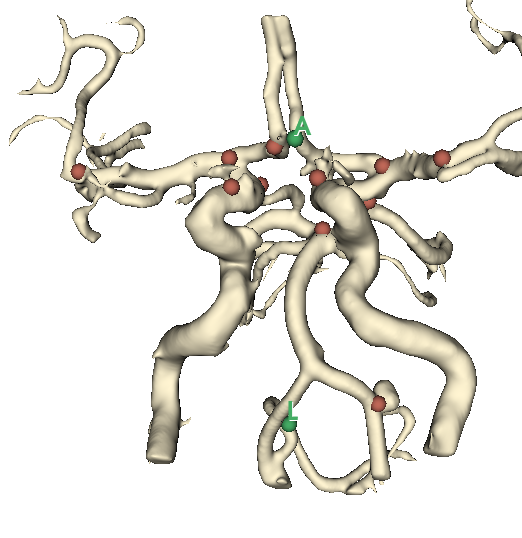}}
		\hfill
		\subfloat[SCN]{%
			\includegraphics[width=0.28\linewidth]{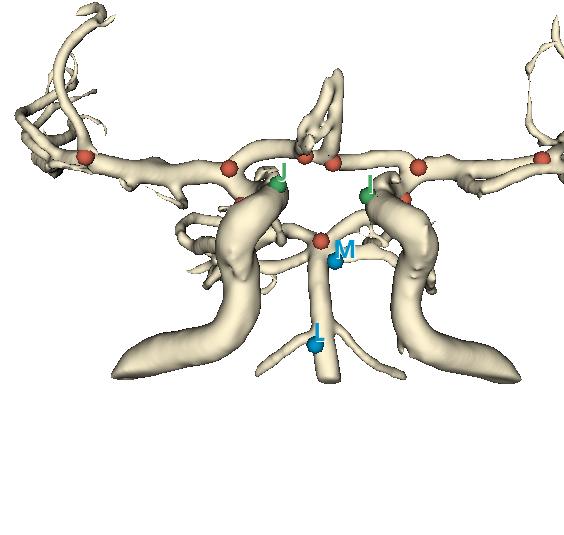}}
		
		\subfloat[FARNet]{%
			\includegraphics[width=0.28\linewidth]{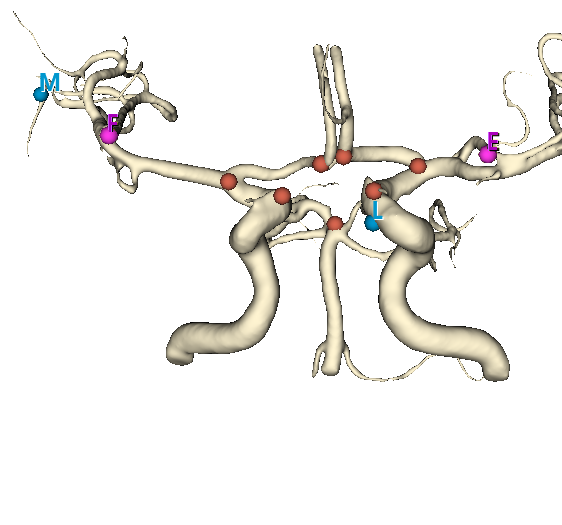}}
		\hfill
		\subfloat[FARNet]{%
			\includegraphics[width=0.28\linewidth]{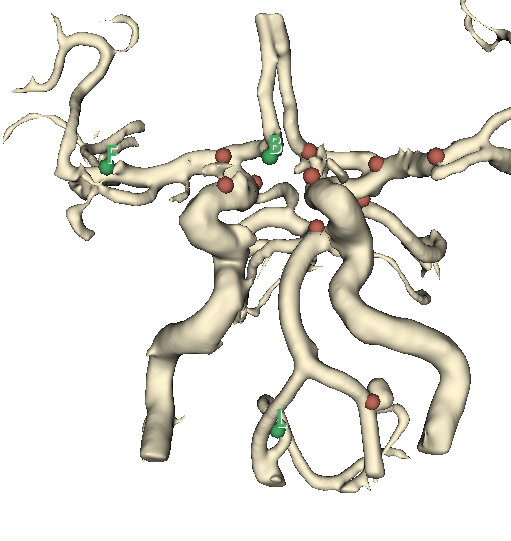}}
		\hfill
		\subfloat[FARNet]{%
			\includegraphics[width=0.28\linewidth]{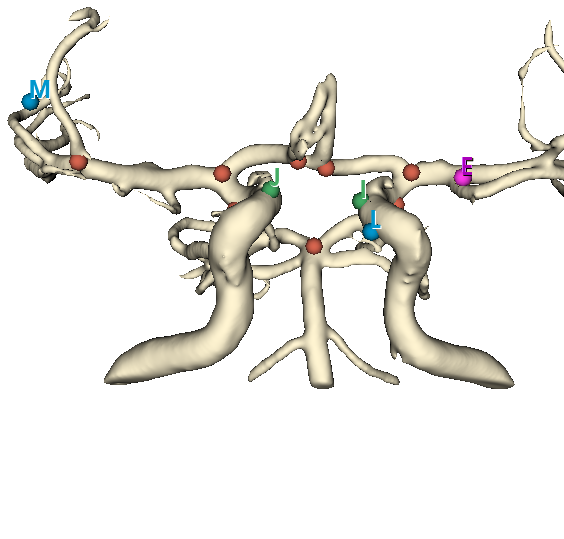}}
		
		\subfloat[OURS]{%
			\includegraphics[width=0.28\linewidth]{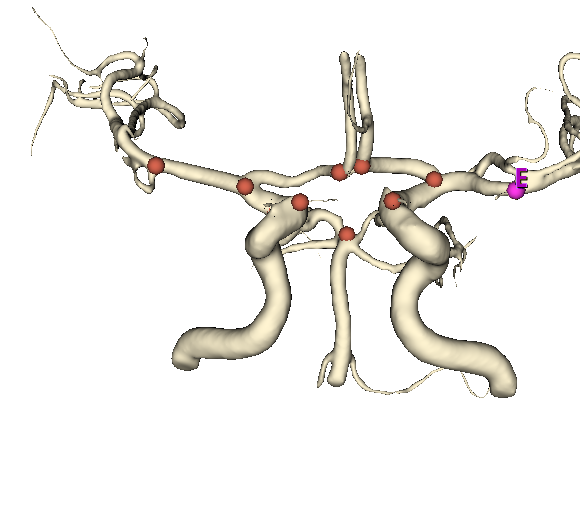}}
		\hfill
		\subfloat[OURS]{%
			\includegraphics[width=0.28\linewidth]{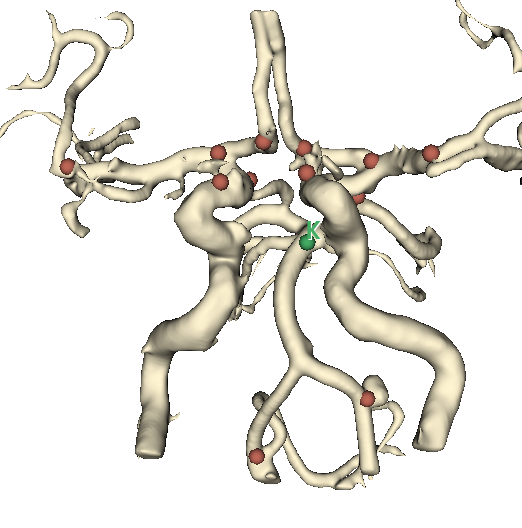}}
		\hfill
		\subfloat[OURS]{%
			\includegraphics[width=0.28\linewidth]{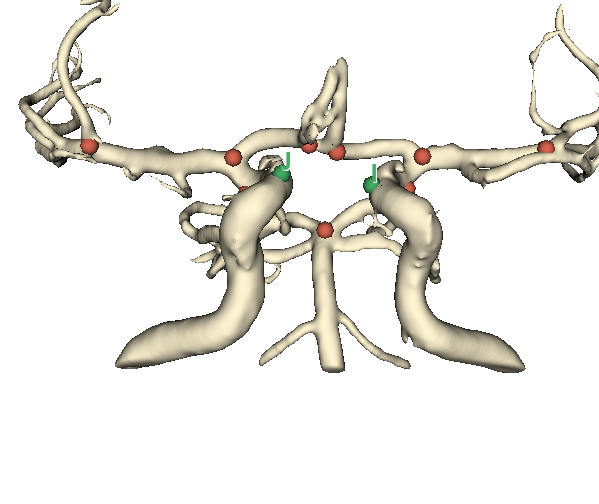}} 
		\caption{Qualitative Comparisons between different methods. Red color refer to GT landmarks or landmarks predicted within $2~mm$. Green color refers to predictions within ($2-5~mm$) distance from GT. Yellow color refers to failure cases (landmarks predicted for wrong bifurcations ($>10~mm$). Blue color refers to false positive predictions.}\label{fig:comp}
	\end{figure*}

	Fig. \ref{fig:bif} depicts the results of the mean radius error (MRE) for each   bifurcation label. Some bifurcations are more difficult to locate than others due to the intricate structure of the branches and the anatomical variations in the Circle of Willis. Notably, the uncertainty associated with division of the middle cerebral artery (MCA) into its M1 and M2 segments  leads to a greater  error (MRE) of $2.4$ and $2.8~mm$ for landmarks E and F, respectively. Furthermore, landmarks L and M exhibit elevated MREs due to the limited availability of data and their inconsistent presence in the dataset. However, for such landmarks, compared to state-of-the-art approaches, our method demonstrates significantly reduced detection error (M with an MRE of $2.12~mm$ using our method compared with $5.68~mm$ and $5.12~mm$ for FARNet and SCN) primarily due to the step 1 detection algorithm that focuses on the most relevant regions of interest (ROI) of labels \#R9 and \#R10. Qualitative comparisons are presented in Fig. \ref{fig:comp}.
	
	\subsubsection{Public MRA dataset} 
	\begin{figure*}[!ht]
		\begin{centering}
			\includegraphics[width=0.72\textwidth]{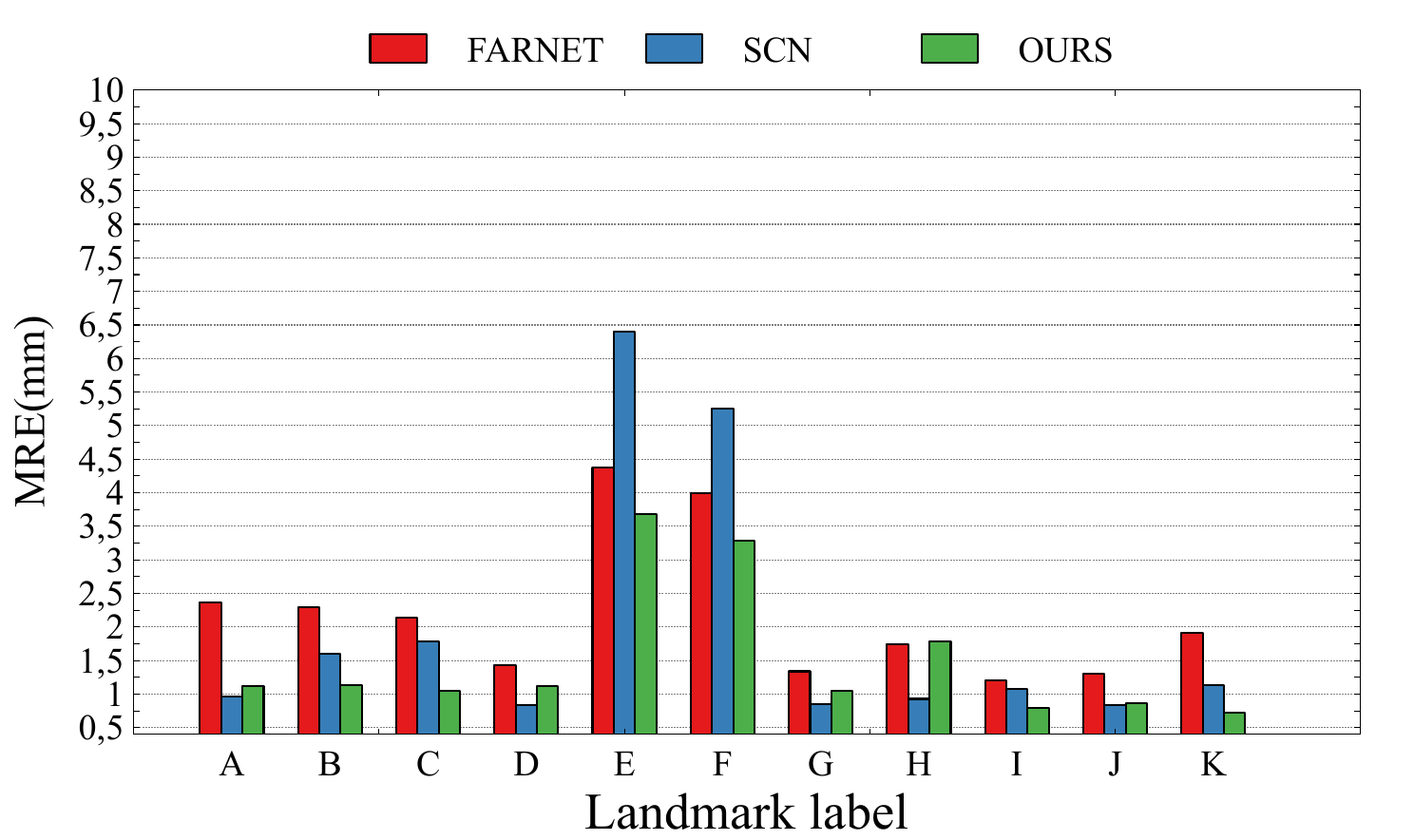}
			\par\end{centering}
		\caption{Mean radial errors(mm) for 11 landmarks labels for the public dataset. \label{fig:bif2}}
	\end{figure*}
	
	\begin{figure}[!ht]
		\begin{centering}
			\includegraphics[width=0.78\columnwidth]{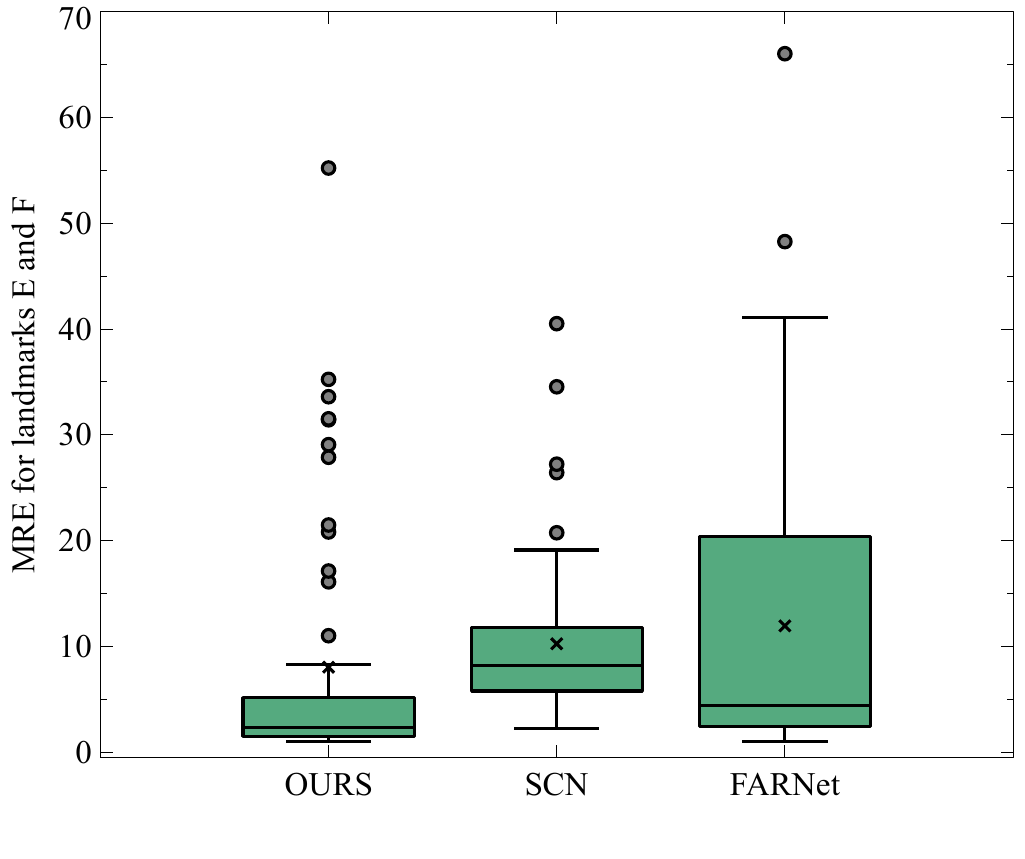}
			\par\end{centering}
		\caption{Distance distribution between the predicted bifurcations and the Ground Truth annotations on Landmarks E and F. \label{fig:EF}}
	\end{figure}
	
	In our investigation using the public MRA dataset, which encompasses a complete set of landmarks, we present the comparative results in Table \ref{tab3}. 
	\begin{table}[!ht]
		\caption{Detection comparison on the public MRA dataset, quantified using MRE and SDR for three distinct neighborhood sizes.  \label{tab3}}
		\centering{}%
		\begin{tabular}{|c|c|c|c|}
			\hline 
			\textbf{Metric/Method} & \textbf{FARNET} & \textbf{SCN} & \textbf{OURS} \tabularnewline
			\hline
			MRE (mm) & 2.21 & 1.72 & \textbf{1.47} \tabularnewline
			\hline 
			SDR($\%$)-3 & 82.49 & 91.91 & \textbf{92.59} \tabularnewline
			\hline 
			SDR($\%$)-4 & 89.89 & 93.26 & \textbf{94.94} \tabularnewline
			\hline 
			SDR($\%$)-5 & \textbf{96.28} & 93.60 & 95.62 \tabularnewline
			\hline
		\end{tabular}
	\end{table}
	
	The results of our method show a median radial error (MRE) of $1.47~mm$, indicating moderate improvements  to the SCN and FARNet approaches, with an increase of $0.24~mm$ and $0.77~mm$ respectively.  A smaller mean relative error (MRE) indicates a higher level of accuracy in determining the true positions of landmarks. This is particularly important for applications with strong requirements on the spatial accuracy. Our approach surpasses the performance of previous methods when it comes to detecting landmarks inside tighter boundaries, specifically for the $3$ and $4~mm$ neighborhood radius, achieving SDRs of $92.59\%$ and $94.94\%$ respectively. When considering the $5~mm$ threshold, even though FARNet achieves the maximum SDR of $96.28\%$, our technique still achieves a competitive performance of $95.62\%$.
	
	As illustrated in Fig.~\ref{fig:bif2}, the detection performance per label demonstrates a pattern consistent with that observed for the In-House dataset (Fig.~\ref{fig:bif}). The bifurcation landmarks E and F exhibit larger detection errors compared to other bifurcations (excluding L and M).
	However, our method outperforms SCN and FARNet in detecting landmarks E and F. To further analyze this, we present boxplots of the mean radial errors for landmarks E and F in Fig.~\ref{fig:EF}. Our method presents the lowest median and whiskers for landmarks E and F. This implies that it generally predicts landmarks that are closer to the true location, demonstrating a good level of accuracy for most landmarks. The  outliers that can be generated by our method represent occasional mistakes which can be attributed to misclassification  occurrences (proximal bounding boxes) in step 1  due to similarities in appearance with the target landmarks of MCA-M1 and M2 branches.
	SCN and FARNet present larger whiskers and fewer outliers compared to OURS. This indicates a wider range of distances, resulting in less consistency. They often have significant difficulty in discriminating between characteristics associated  within the tree structure along the MCA.

	\begin{figure*}
		\centering
		\subfloat[Detection of Landmark B within $2~mm$ with the presence of a big aneurysm (green shaded volume). \label{fig:im1}]{%
			\includegraphics[width=0.32\linewidth]{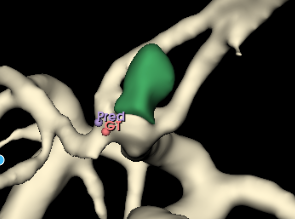}}
		\hfill
		\subfloat[Detection of Landmark E within $2~mm$ with the presence of a small aneurysm.  \label{fig:im2}]{%
			\includegraphics[width=0.31\linewidth]{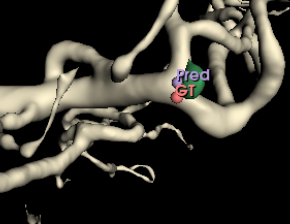}}
		\hfill
		\subfloat[Misdetection of Landmark F. It is a failure case related to MCA arterial tree and not  related to the presence of an aneurysm.   \label{fig:im3}]{%
			\includegraphics[width=0.32\linewidth]{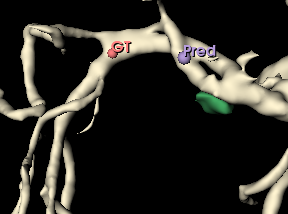}}
		
		\caption{Qualitative results on landmark detection with the presence of aneurysms}\label{detection_aneu}
	\end{figure*}
	
	\subsection{Detection of aneurysm bearing bifurcations}
	To better emphasize the impact of aneurysm presence on the detection of a landmark, we gathered data from the ICAN database comprising 134 MRA-TOF images, each one presenting an aneurysm, 'with various volumes going from $2~mm^3$ to $1050~mm^3$. We annotated the landmark coordinates and labeled the specific locations where the aneurysms occurred. The distribution of these aneurysms is detailed in Table \ref{tab4}.
	We calculated the detection rates of landmarks within five different distance thresholds: $2~mm$, $3~mm$, $4~mm$, $5~mm$, and $10~mm$. The findings are as follows: 60$\%$ of landmarks were detected within a $2~mm$ radius, 73$\%$ within $3~mm$, 87$\%$ within $4~mm$, 90$\%$ within $5~mm$, and 97$\%$ within a $10~mm$ radius.

	\begin{table}[!ht]
		\caption{Distribution of aneurysm for each bifurcation label. The landmarks H and K were missed from the dataset\label{tab4}}
		
		\centering{}%
		\begin{tabular}{|c|c|c|c|c|c|c|c|c|c|}
			\hline 
			\textbf{Bifurcation} & A & B & C & D & E &F & G&I&J\tabularnewline
			\hline 
			\textbf{\# of ICAs} & 7& 11 & 6 & 4 & 19 & 39 & 3&28&17\tabularnewline
			\hline 
		\end{tabular}
	\end{table}
	In a comparative analysis of landmark detection rates with and without the presence of aneurysms, we observed a similar SDR of approximately 95$\%$ for a $5~mm$ threshold. This suggests that the presence of aneurysms does not significantly impact the accuracy of  detection. However, it is noteworthy that four landmarks were predicted within a distance  greater than $10~mm$.
	The failure cases are linked to bifurcations E and F and are caused by the inaccurate prediction, by nnDetction network, of the Region of Interest (ROI) of labels \#R4 and \#R5 due to the resemblance in the MCA branching patterns. Fig \ref{detection_aneu} shows qualitative results of landmark detection with the presence of aneurysms.

	\section{Conclusion}
	\label{sec:Conclusion}
	In this paper, we propose a two-steps deep learning-based framework for landmark detection of bifurcations along the circle of Willis.
	Initially, the challenge is framed as a task of object classification paired with bounding box regression. Each landmark is enclosed in a fixed-size bounding box that is positioned centrally around the landmark. Using a 3D nnDetection network, we detect and classify potential regions of interest (ROIs), assigning each ROI a specific label and score. In the second step, the precise locations of the landmarks are determined using heatmap regression via a UNet model.
	The experimental results demonstrate that our method achieves location accuracy that is comparable to the leading heatmap regression algorithms ~\cite{payer2019integrating,ao2023feature}, as seen by the minimal mean radial error (MRE). Our method leads to a reduction in misidentification errors of similar bifurcations seen in the proximal branches (landmarks E and F). In addition, our two-step approach is specifically designed to handle datasets that have a  varying number of landmarks. This differentiates our proposed technique from heatmap regression methods, which are not designed for such an eventuality. Our approach is highly beneficial, as evidenced by landmarks L and M, as it reduces the chances of false positives.
	
	In order to enhance performance, particularly in accurately identifying similar bifurcations within the MCA branch, it would be beneficial to augment the training dataset with a more diverse assortment of MCA landmarks. Expanding the application of the detection model to a broader spectrum of anatomical differences should improve its capacity to precisely recognize and distinguish closely situated and visually similar bifurcations, hence boosting the overall accuracy of the detection process.

	\bibliographystyle{IEEEtranS}
	\bibliography{refs}

\end{document}